\documentclass[final]{cvpr}

\usepackage{times}
\usepackage{epsfig}
\usepackage{graphicx}
\usepackage{amsmath}
\usepackage{amssymb}

\usepackage{multirow}
\usepackage{caption}

\newcommand{\indep}{\perp \!\!\! \perp}

\usepackage[pagebackref=true,breaklinks=true,colorlinks,bookmarks=false]{hyperref}

\def\cvprPaperID{6228} 
\def\confYear{CVPR 2021}

\begin{document}

\title{Probabilistic 3D Human Shape and Pose Estimation from Multiple Unconstrained Images in the Wild}

\author{Akash Sengupta\\
University of Cambridge\\
{\tt\small as2562@cam.ac.uk}
\and
Ignas Budvytis\\
University of Cambridge\\
{\tt\small ib255@cam.ac.uk}
\and 
Roberto Cipolla\\
University of Cambridge\\
{\tt\small rc10001@cam.ac.uk}
}

\maketitle

\begin{abstract}
   This paper addresses the problem of 3D human body shape and pose estimation from RGB images. Recent progress in this field has focused on single images, video or multi-view images as inputs. In contrast, we propose a new task: shape and pose estimation from a group of multiple images of a human subject, without constraints on subject pose, camera viewpoint or background conditions between images in the group. Our solution to this task predicts distributions over SMPL body shape and pose parameters conditioned on the input images in the group. We probabilistically combine predicted body shape distributions from each image to obtain a final multi-image shape prediction. We show that the additional body shape information present in multi-image input groups improves 3D human shape estimation metrics compared to single-image inputs on the SSP-3D dataset and a private dataset of tape-measured humans. In addition, predicting distributions over 3D bodies allows us to quantify pose prediction uncertainty, which is useful when faced with challenging input images with significant occlusion. Our method demonstrates meaningful pose uncertainty on the 3DPW dataset and is competitive with the state-of-the-art in terms of pose estimation metrics.
\end{abstract}
\vspace{-0.4cm}

\section{Introduction}

\begin{figure}[t]
    \centering
    \includegraphics[width=1.0\linewidth]{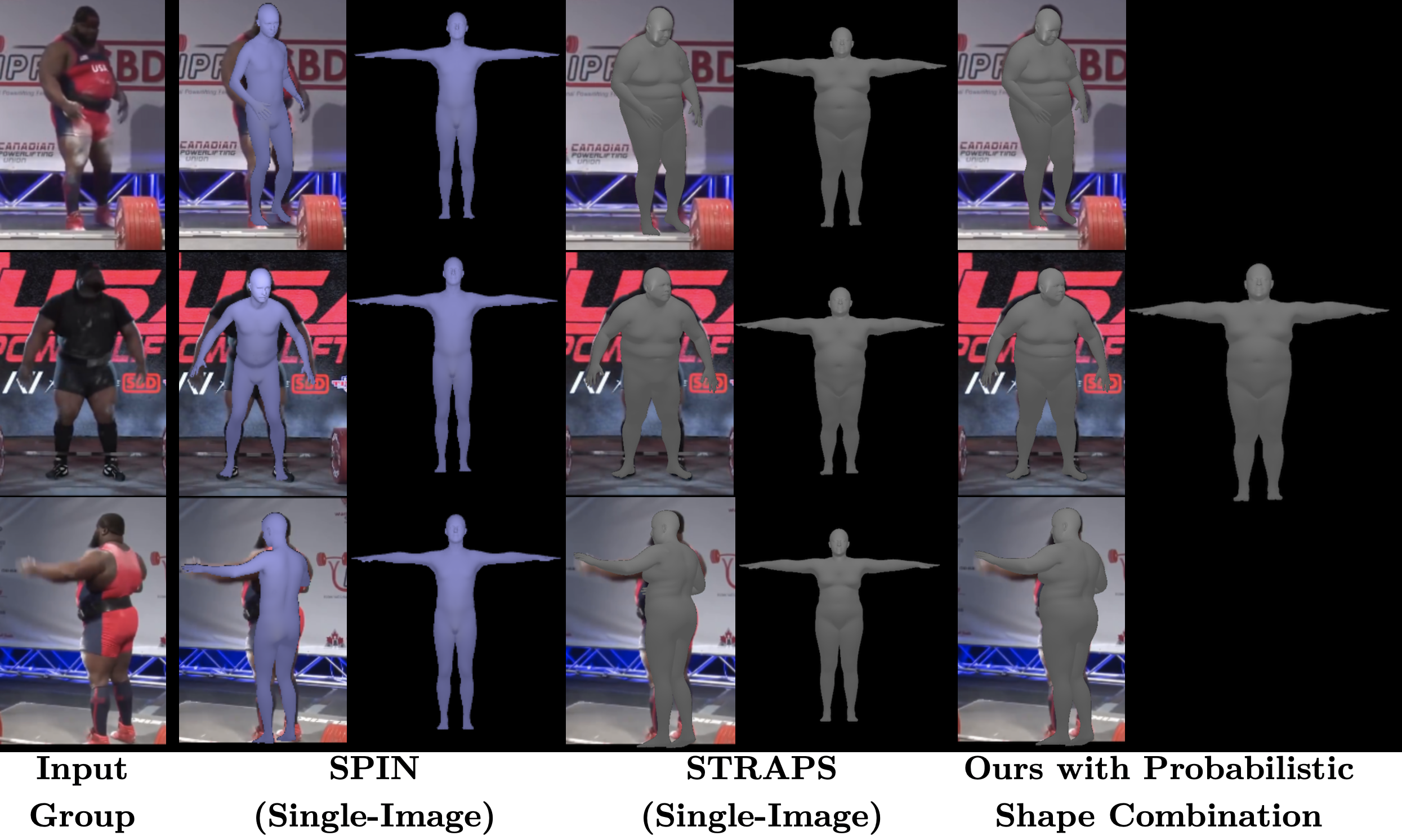}
    \caption{Example shape and pose predictions from a group of input images. Probabilistic shape combination results in a more accurate body shape estimate than both individual single-image predictions (visualised here from SPIN \cite{kolotouros2019spin} and STRAPS \cite{STRAPS2020BMVC}) and naively-averaged single-image predictions, as our experiments show in Section \ref{sec:experimental_results}.}
    \label{fig:intro}
    \vspace{-0.4cm}
\end{figure}

3D human body shape and pose estimation from RGB images is a challenging problem with potential applications in augmented and virtual reality, healthcare and fitness technology and virtual retail. Recent solutions have focused on three types of inputs: i) single images \cite{Bogo:ECCV:2016, tan2017,hmrKanazawa17,kolotouros2019spin,kolotouros2019cmr,zhang2019danet,Moon_2020_ECCV_I2L-MeshNet,STRAPS2020BMVC,omran2018nbf,pavlakos2018humanshape,varol18_bodynet}, ii) video \cite{kocabas2019vibe,humanMotionKanazawa19,sun2019dsd-satn, pavlakos2019texturepose, MuVS:3DV:2017} with temporal constraints on pose, camera viewpoint and background conditions and iii) multi-view images \cite{liang2019samv,smith20193dfromsilhouettes} with a fixed subject pose captured from multiple viewpoints. In contrast, we aim to estimate 3D body shape and pose from a group of images of the same human subject without any constraints on the subject's pose, camera viewpoint or background conditions between the images, as illustrated in Figure \ref{fig:intro}. This task is motivated by the intuition that multiple images of the same subject should contain additional visual information about their body shape compared to a single image, regardless of whether the subject's pose or surrounding environment change between images. A suitable shape and pose estimator should leverage this information to improve shape prediction accuracy over single-image methods.

\begin{figure*}[t]
    \centering
    \includegraphics[width=0.9\linewidth]{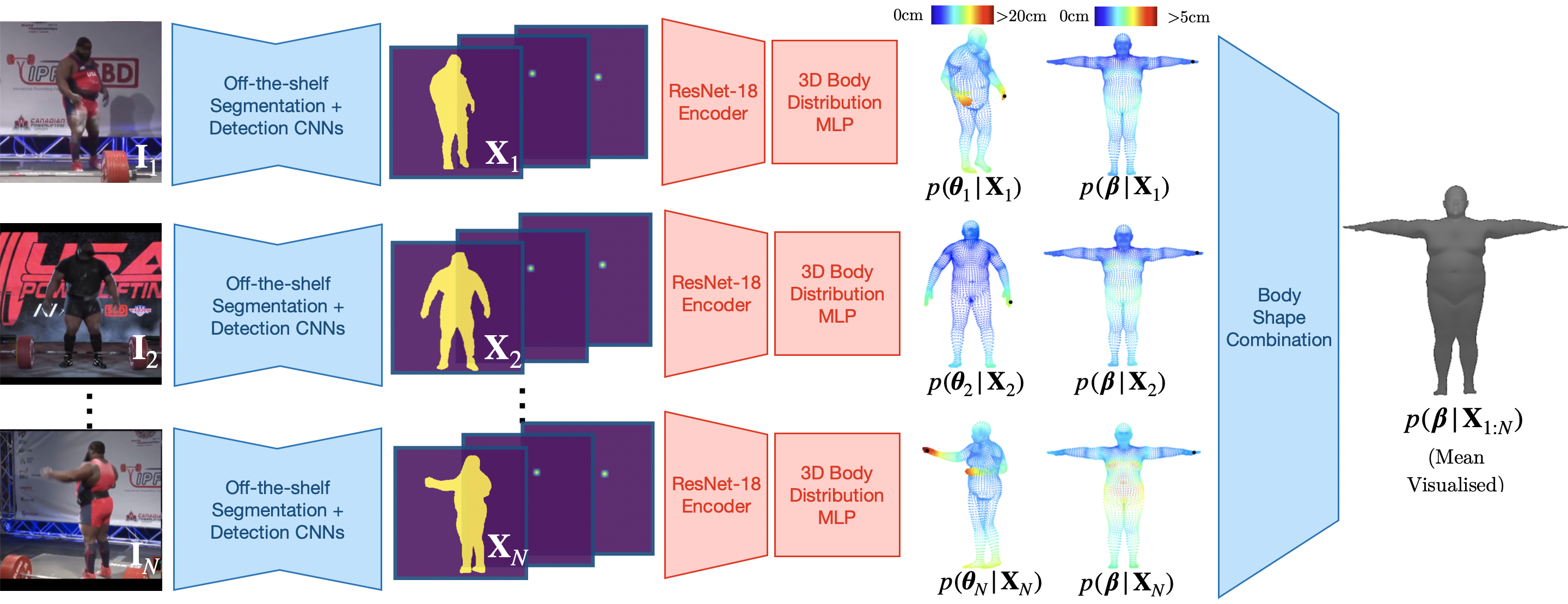}
    \caption{Overview of our shape and pose distribution prediction network. Each image $\mathbf{I}_n$ in the input group is converted into a silhouette and joint proxy representation $\mathbf{X}_n$, which is passed through a distribution prediction network to obtain multivariate distributions over SMPL \cite{SMPL:2015} shape and pose parameters, $\boldsymbol{\beta}$ and $\boldsymbol{\theta}_n$, conditioned on the input. Shape distributions from each individual input are probabilistically combined to form a multi-input shape distribution. The encoder and distribution MLP are trained using randomly-generated synthetic data \cite{STRAPS2020BMVC}. The per-vertex uncertainty visualisations (in cm) are obtained by sampling SMPL parameters from the predicted distributions, computing the SMPL vertex mesh for each sample and determining the average Euclidean distance from the mean for each vertex. Black dots indicate left hands.}
    \label{fig:method}
    \vspace{-0.3cm}
\end{figure*}

We present a probabilistic body shape and pose estimation method from a group of unconstrained images of the same subject. Inference occurs in three stages (see Figure \ref{fig:method}). First, we predict a proxy representation from each input image in the group, consisting of the subject's silhouette and 2D joint location heatmaps, using off-the-shelf segmentation and 2D keypoint detection CNNs \cite{he2017maskrcnn, Guler2018DensePose, kirillov2019pointrend, wu2019detectron2}. Then, each proxy representation is passed through a 3D distribution prediction network that outputs a probability distribution over SMPL \cite{SMPL:2015} body shape and pose parameters conditioned on the input representation. Lastly, body shape distributions from each input image are probabilistically combined to procure a final shape prediction. This yields a better estimate of the subject's body shape than current single-image body shape and pose estimators \cite{hmrKanazawa17,kolotouros2019spin,kolotouros2019cmr,zhang2019danet,STRAPS2020BMVC}, which may be inaccurate or inconsistent, as shown in Figure \ref{fig:intro}.

Moreover, most single-image body model parameter regressors \cite{hmrKanazawa17,kolotouros2019spin,omran2018nbf,zhang2019danet,STRAPS2020BMVC,Xu_2019_ICCV, georgakis2020hkmr} do not consider the uncertainty associated with each pose parameter estimate. If certain body parts are occluded or out-of-frame in the input image, the estimator can only guess about the pose parameters corresponding to these body parts. Such situations further motivate our approach of predicting a distribution over body pose, since the variance of the distribution quantifies the uncertainty associated with each pose parameter prediction, as shown in Figures \ref{fig:synthtic_data_example_predictions} and \ref{fig:3dpw_uncertainty}.

Training body model parameter regressors to accurately predict body shape is challenging due to the lack of suitable training datasets of in-the-wild images paired with \textit{accurate and diverse} body shape labels. Collecting such data is practically difficult, particularly for our proposed task of shape estimation from a group of unconstrained images. Recent works \cite{STRAPS2020BMVC, smith20193dfromsilhouettes, varol17_surreal} propose using synthetic input-label pairs to overcome the lack of suitable training datasets. We adopt the same synthetic training approach as STRAPS \cite{STRAPS2020BMVC} to train our 3D distribution prediction network, but extend the data augmentations used to bridge the gap between synthetic and real inputs. In particular, our synthetic training data better models occluded and out-of-frame body parts in silhouettes and joints such that the domain gap to real occluded data is smaller. This allows our method to estimate pose prediction uncertainty and also results in improved single-input pose prediction metrics on challenging evaluation datasets, such as 3DPW \cite{vonMarcard2018}.

In summary, our main contributions are as follows:
\begin{itemize}
    \item We propose a novel task: predicting body shape from a group of images of the same human subject, without imposing any constraints on subject pose, camera viewpoint or backgrounds between the images.
    \item We present a solution to the proposed task which predicts a distribution over 3D human body shape and pose parameters conditioned on the input images in the group. Body shape distributions from each image are probabilistically combined to yield a final body shape estimate which leverages multi-image shape information, resulting in a more accurate body shape estimate compared to single-input methods.
    \item To the best of our knowledge, our method is the first to output uncertainties alongside associated SMPL \cite{SMPL:2015} shape and pose parameter predictions, which are shown to be useful when input images contain occluded or out-of-frame body parts.
    \item We extend the synthetic training framework introduced by \cite{STRAPS2020BMVC} to better model occlusion and missing body parts, allowing our synthetically-trained distribution prediction neural network to yield better 3D shape and pose metrics.
\end{itemize}

\vspace{-0.3cm}
\section{Related Work}

This section discusses recent approaches to 3D human shape and pose estimation from single images, multi-view images and video.

\noindent \textbf{Single-image shape and pose estimation} methods can be classified into 2 categories: optimisation-based and learning-based. Optimisation-based methods fit a parametric 3D body model \cite{SMPL:2015, SMPL-X:2019, Anguelov05scape:shape, Joo_2018_CVPR_total_capture} to 2D observations (e.g. 2D joints \cite{Bogo:ECCV:2016, SMPL-X:2019}, surface landmarks \cite{Lassner:UP:2017}, silhouettes \cite{Lassner:UP:2017} or part segmentations \cite{Zanfir_2018_CVPR}) via optimisation. They can accurately estimate 3D poses without requiring expensive 3D-labelled datasets, however they are susceptible to poor initialisation and tend to be slow at test-time.

Learning-based methods can be further classified as model-free or model-based. Model-free approaches directly predict a 3D body representation from an image, such as a voxel occupancy grid \cite{varol18_bodynet}, vertex mesh \cite{kolotouros2019cmr, Moon_2020_ECCV_I2L-MeshNet, Zeng_2020_CVPR_mesh_dense} or implicit surface representation \cite{saito2020pifuhd}. Model-based approaches \cite{hmrKanazawa17, zhang2019danet, omran2018nbf, pavlakos2018humanshape, tan2017, georgakis2020hkmr, Guler_2019_CVPR_holopose} predict the parameters of a 3D body model \cite{SMPL:2015, SMPL-X:2019, Anguelov05scape:shape, Joo_2018_CVPR_total_capture}, which provides a useful prior over human body shape. Several methods \cite{hmrKanazawa17, kolotouros2019cmr, Guler_2019_CVPR_holopose, Xu_2019_ICCV} overcome the scarcity of in-the-wild 3D-labelled training data by incorporating weak supervision with datasets of labelled 2D keypoints. \cite{kolotouros2019spin} extends this further by integrating optimisation into the model training loop to lift 2D keypoint labels to self-improving 3D pose and shape labels. Such approaches predict impressive 3D poses but fail to predict accurate body shapes (particularly for non-average humans) since 2D keypoints do not densely inform shape. Recently, \cite{STRAPS2020BMVC} used random synthetic training data to overcome data scarcity and demonstrated improved shape predictions.

\noindent \textbf{Video shape and pose estimation} methods may be classified similarly to their single-image counterparts. Optimisation-based video methods \cite{Arnab_CVPR_2019, MuVS:3DV:2017, alldieck2017optical} extend single-image optimisation over time, while learning-based video methods \cite{humanMotionKanazawa19, kocabas2019vibe, sun2019dsd-satn, NIPS2017_7108, pavlakos2019texturepose} modify single-image predictors to take sequences of frames as inputs. However, video inputs allow these methods to enforce consistent body shapes and smooth motions across frames, e.g. using motion discriminators \cite{movi2020}, optical flow \cite{alldieck2017optical, NIPS2017_7108}, or texture consistency \cite{pavlakos2019texturepose}. Learning-based video methods also overcome 3D data scarcity by incorporating weak 2D supervision \cite{kocabas2019vibe, humanMotionKanazawa19}, or with self-supervision enforcing visual consistency between frames \cite{NIPS2017_7108, pavlakos2019texturepose, alldieck2017optical}. Nevertheless, current methods are unable to predict accurate body shapes, particularly for non-average humans.

\noindent \textbf{Multi-view shape and pose estimation.} \cite{liang2019samv} extends the iterative regressor of \cite{hmrKanazawa17} to predict body model parameters from multiple input images of the same subject in a fixed pose, captured from varying camera angles. They use synthetic data to overcome data scarcity, resulting in more accurate body shape estimates, particularly under clothing. \cite{smith20193dfromsilhouettes} uses synthetic data to learn to predict body model parameters from A-pose silhouettes.

Contrary to the above approaches, our method estimates shape and pose from a group of images without \textit{any} temporal or absolute constraints on the subject's pose, camera viewpoint or background between images.

\noindent \textbf{Shape and pose distribution estimation.} While substantial progress has been made in predicting probability distributions using neural networks \cite{baum1988superviseddistribution, nix1994icnn, Bishop94mixturedensity, kendall2017whatuncertainties, rupprecht2017learning, prokudin2018deepdirectstat, li2019posemdn, mohlin2020matrixfisher}, prediction of distributions over 3D human shape \textit{and} pose is still under-explored. Recently, \cite{Moon_2020_ECCV_I2L-MeshNet} used lixel-based 1D heatmaps to quantify uncertainty in predicted 3D human mesh vertex locations. \cite{biggs2020multibodies} predicted a categorical distribution over multiple SMPL hypotheses given an ambiguous image. In contrast, we aim to explicitly output separable uncertainties per predicted pose and shape parameter, since the shape uncertainties, specifically, are used for shape prediction from multiple unconstrained images.

\section{Method}

This section provides a brief overview of the SMPL parametric human body model  \cite{SMPL:2015}, presents our three-stage method for probabilistic body shape and pose estimation from a group of unconstrained images of the same human subject (illustrated in Figure \ref{fig:method}) and finally discusses the synthetic training framework and loss functions used.

\subsection{SMPL model}

SMPL \cite{SMPL:2015} provides a differentiable function $\mathcal{M}(\boldsymbol{\theta}, \boldsymbol{\beta}, \boldsymbol{\gamma})$ which takes pose parameters $\boldsymbol{\theta}$, global rotation $\boldsymbol{\gamma}$ and identity-dependent body shape parameters $\boldsymbol{\beta}$ as inputs and outputs a vertex mesh $\mathbf{V} \in \mathbb{R}^{6890 \times 3}$. $\boldsymbol{\theta} \in \mathbb{R}^{69}$ and $\boldsymbol{\gamma} \in \mathbb{R}^3$ represent axis-angle rotation vectors for 23 SMPL body joints and the root joint respectively. $\boldsymbol{\beta} \in \mathbb{R}^{10}$ represents coefficients of a PCA body shape basis. Given the vertex mesh $\mathbf{V}$, 3D joint locations may be obtained using a linear regressor, $\mathbf{J}^{\text{3D}} = \mathcal{J}\mathbf{V}$ where $\mathcal{J} \in \mathbb{R}^{L \times 6890}$ is a regression matrix for $L$ joints of interest.

\subsection{Proxy representation computation}

Given a group of $N$ RGB input images $\{\mathbf{I}_n\}_{n=1}^N$ of the same subject, we first compute proxy representations $\{\mathbf{X}_n\}_{n=1}^N$. DensePose \cite{Guler2018DensePose} is used to obtain body part segmentations, which are converted into silhouettes. Keypoint-RCNN from Detectron2 \cite{he2017maskrcnn, wu2019detectron2} is used to obtain 2D joint locations, which are converted into Gaussian heatmaps, and associated confidence scores. Heatmaps corresponding to joint detections with confidence scores less than a threshold $t=0.025$ are set to 0. Thresholding is essential for modelling uncertainty in 3D pose predictions as it typically removes invisible 2D joints from the input representations. The predicted silhouette and joint heatmaps from each image are stacked along the channel dimension to form each proxy representation $\mathbf{X}_n \in \mathbb{R}^{H \times W \times (L+1)}$.

The use of silhouette and joint heatmap representations as inputs instead of RGB images is inspired by \cite{pavlakos2018humanshape, STRAPS2020BMVC} and allows us to train our distribution prediction network using a simple synthetic training framework (see Section \ref{subsec:network_training}), overcoming the lack of shape diversity in current datasets. We follow \cite{STRAPS2020BMVC} and use simple silhouettes and 2D joint heatmaps as our proxy representation, instead of more complex alternatives (e.g. part segmentations or IUV maps), since this leads to a smaller synthetic-to-real domain gap which is more readily bridged by data augmentation \cite{STRAPS2020BMVC}.

\subsection{Body shape and pose distribution prediction}

We aim to estimate probability distributions \cite{nix1994icnn} over SMPL pose parameters $\{\boldsymbol{\theta}_n\}_{n=1}^N$ (which are free to change between inputs) and the subject's identity-dependent shape $\boldsymbol{\beta}$, both conditional upon $\{\mathbf{X}_n\}_{n=1}^N$. We assume simple multivariate Gaussian distributions
\begin{equation}
\begin{aligned}
p(\boldsymbol{\theta}_n | {\mathbf{X}_n}) & = \mathcal{N}(\boldsymbol{\theta}_n ; \boldsymbol{\mu}_\theta({\mathbf{X}_n}), \boldsymbol{\Sigma}_\theta ({\mathbf{X}_n}))\\
p(\boldsymbol{\beta} | {\mathbf{X}_n}) & = \mathcal{N}(\boldsymbol{\beta} ; \boldsymbol{\mu}_\beta({\mathbf{X}_n}), \boldsymbol{\Sigma}_\beta ({\mathbf{X}_n})).
\end{aligned}
\label{eqn:pose_shape_distributions}
\vspace{-0.1cm}
\end{equation}
Covariance matrices are constrained to be diagonal, i.e. $\boldsymbol{\Sigma}_\theta (\mathbf{X}_n) = \text{diag}(\boldsymbol{\sigma}^2_\theta (\mathbf{X}_n))$ and $\boldsymbol{\Sigma}_\beta (\mathbf{X}_n) = \text{diag}(\boldsymbol{\sigma}^2_\beta(\mathbf{X}_n))$. Formally, $\boldsymbol{\sigma}^2_\theta(\mathbf{X}_n)$ and $\boldsymbol{\sigma}^2_\beta(\mathbf{X}_n)$ represent estimates of the heteroscedastic aleatoric uncertainty \cite{derkiureghian2009aleatoric_epistemic, kendall2017whatuncertainties} in the SMPL parameters explaining the input observations $\mathbf{X}_n$, which arises particularly due to occlusion.

We also predict deterministic estimates of the global rotations $\{\boldsymbol{\gamma}_n\}_{n=1}^N$ and weak-perspective camera parameters $\{\mathbf{c}_n\}_{n=1}^N$, where $\mathbf{c}_n = [s_n, t^x_n, t^y_n]$ representing scale and $xy$ translation respectively. Global rotation and camera parameters are unconstrained across images.

Hence, we require a function mapping each input proxy representation $\mathbf{X}_n$ to the desired set of outputs $\mathbf{Y}(\mathbf{X}_n) = \{\boldsymbol{\mu}_\theta, \boldsymbol{\mu}_\beta, \boldsymbol{\sigma}^2_\theta, \boldsymbol{\sigma}^2_\beta, \boldsymbol{\gamma}_n, \mathbf{c}_n\}$. This function is represented using a deep neural network $f$ with learnable weights $\mathbf{W}$:
\begin{equation}
\mathbf{Y} = f(\textbf{X}_n; \mathbf{W}).
\label{eqn:neural_network}
\vspace{-0.1cm}
\end{equation}
$f$ consists of a convolutional encoder for feature extraction followed by a simple multi-layer perceptron that predicts the set of outputs $\mathbf{Y}$, as illustrated in Figure \ref{fig:method}. The network training procedure is detailed in Section \ref{subsec:network_training}.

\subsection{Body shape combination}
\label{subsec:body_shape_combination}

We combine the conditional body shape distributions output by $f$ given each individual input, $p(\boldsymbol{\beta} | \mathbf{X}_n)$ for $n = 1, ..., N$, into a final distribution $p(\boldsymbol{\beta} | \{\mathbf{X}_n\}_{n=1}^N)$ that aggregates shape information across the input group. Formally,
\begin{equation}
p(\boldsymbol{\beta} | \{\mathbf{X}_n\}_{n=1}^N) 
\propto \prod_{n=1}^N p(\boldsymbol{\beta} | \mathbf{X}_n)
\label{eqn:probabilistic_shape_combination}
\vspace{-0.1cm}
\end{equation}
which follows from the conditional independence assumption $(\mathbf{X}_i \indep \mathbf{X}_j) | \boldsymbol{\beta}$ for $i,j \in \{1,...,N\}$ and $i \neq j$. This is justifiable since we do not impose any relationship between the subject's pose or camera viewpoint across inputs - only the body shape is fixed. Further details are in the supp. material. Since the product of Gaussians is an un-normalised Gaussian, $p(\boldsymbol{\beta} | \{\mathbf{X}_n\}_{n=1}^N) \propto \mathcal{N}(\boldsymbol{\beta}; \mathbf{m}, \mathbf{S})$ where
\begin{equation}
\begin{aligned}
\mathbf{S} &= \bigg(\sum_{n=1}^N \boldsymbol{\Sigma}^{-1}_\beta ({\mathbf{X}_n})\bigg)^{-1}\\
\mathbf{m} &= \mathbf{S}\bigg(\sum_{n=1}^N \boldsymbol{\Sigma}^{-1}_\beta({\mathbf{X}_n})\boldsymbol{\mu}_\beta({\mathbf{X}_n})\bigg).
\end{aligned}
\label{eqn:combined_shape_mean_var}
\vspace{-0.2cm}
\end{equation}
The combined mean $\mathbf{m}$ is a final point estimate of the subject's body shape from the input group $\{\mathbf{X}_n\}_{n=1}^N$.

\vspace{-0.05cm}
\subsection{Network training}
\label{subsec:network_training}

\begin{figure}[t]
    \centering
    \includegraphics[width=0.8\linewidth]{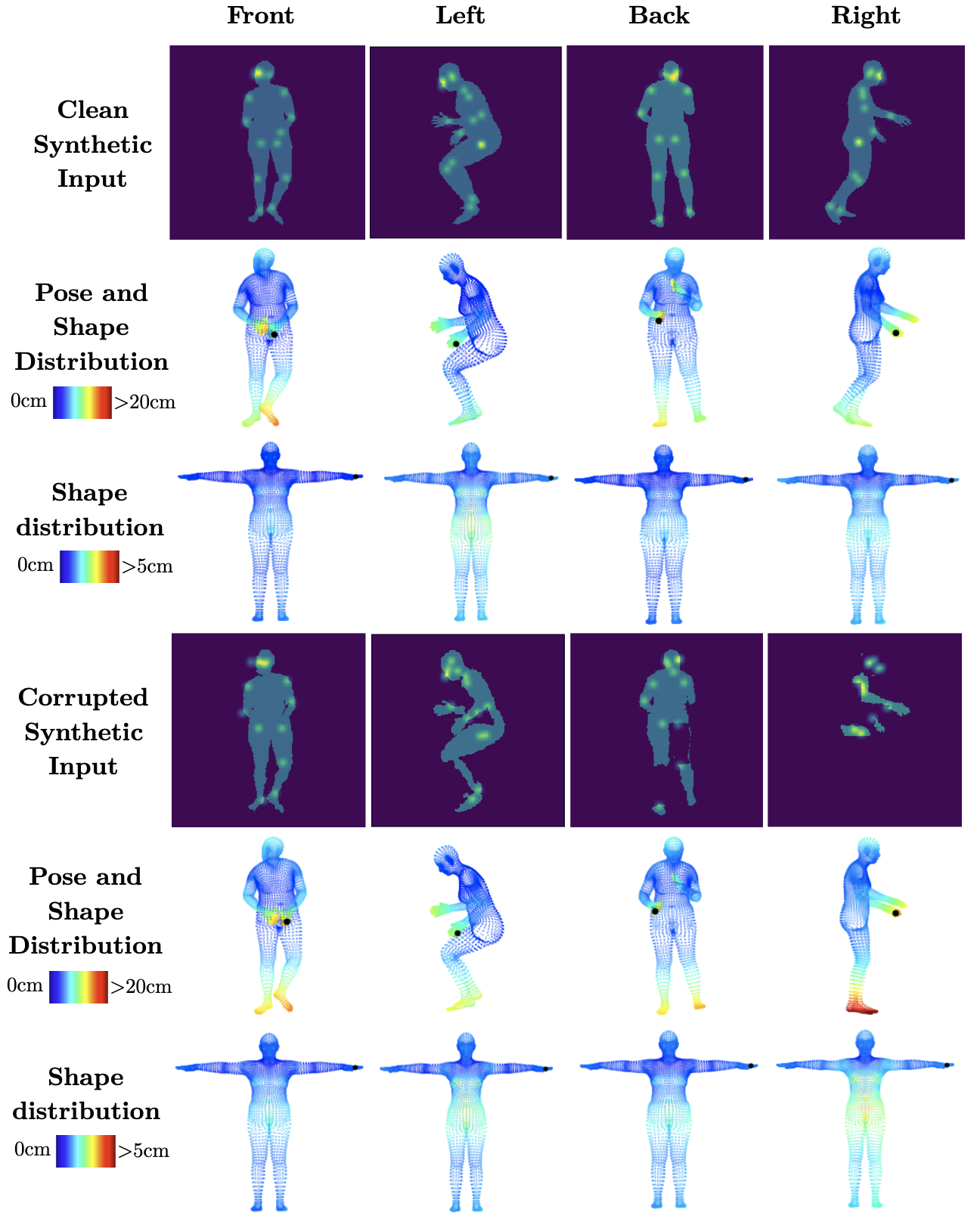}
    \caption{Clean and corrupted versions of an example group of inputs from our synthetic evaluation dataset, and corresponding (single-image) shape and pose distribution predictions. Black dots indicate left hands.}
    \label{fig:synthtic_data_example_predictions}
    \vspace{-0.3cm}
\end{figure}

\begin{figure}[t]
    \centering
    \includegraphics[width=0.82\linewidth]{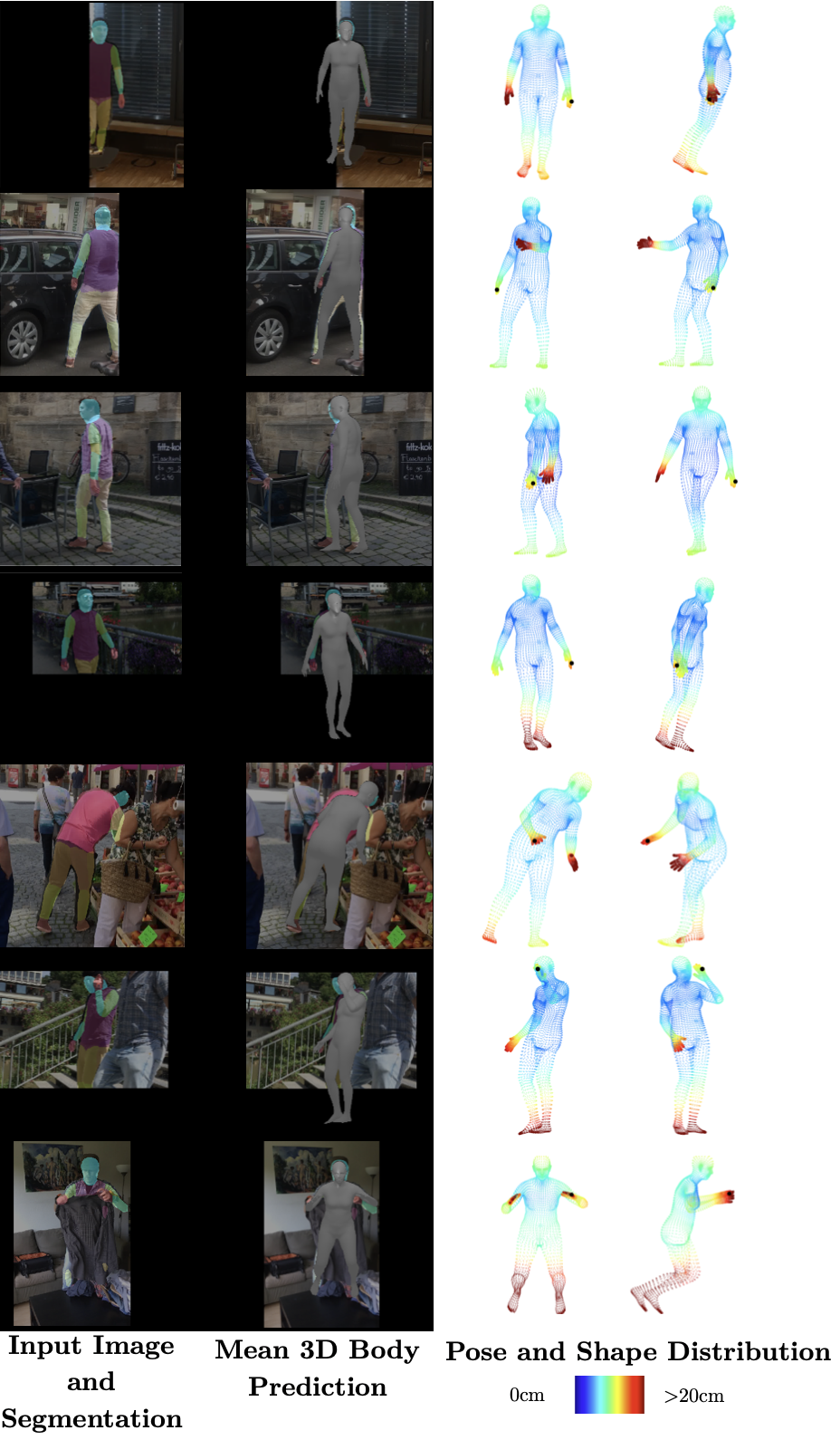}
    \caption{Predictions on single images from 3DPW \cite{vonMarcard2018}. 3D locations of invisible parts are uncertain due to large predicted variances for the corresponding pose parameters.}
    \label{fig:3dpw_uncertainty}
    \vspace{-0.4cm}
\end{figure}

\begin{figure}[t]
    \centering
    \includegraphics[width=1.0\linewidth]{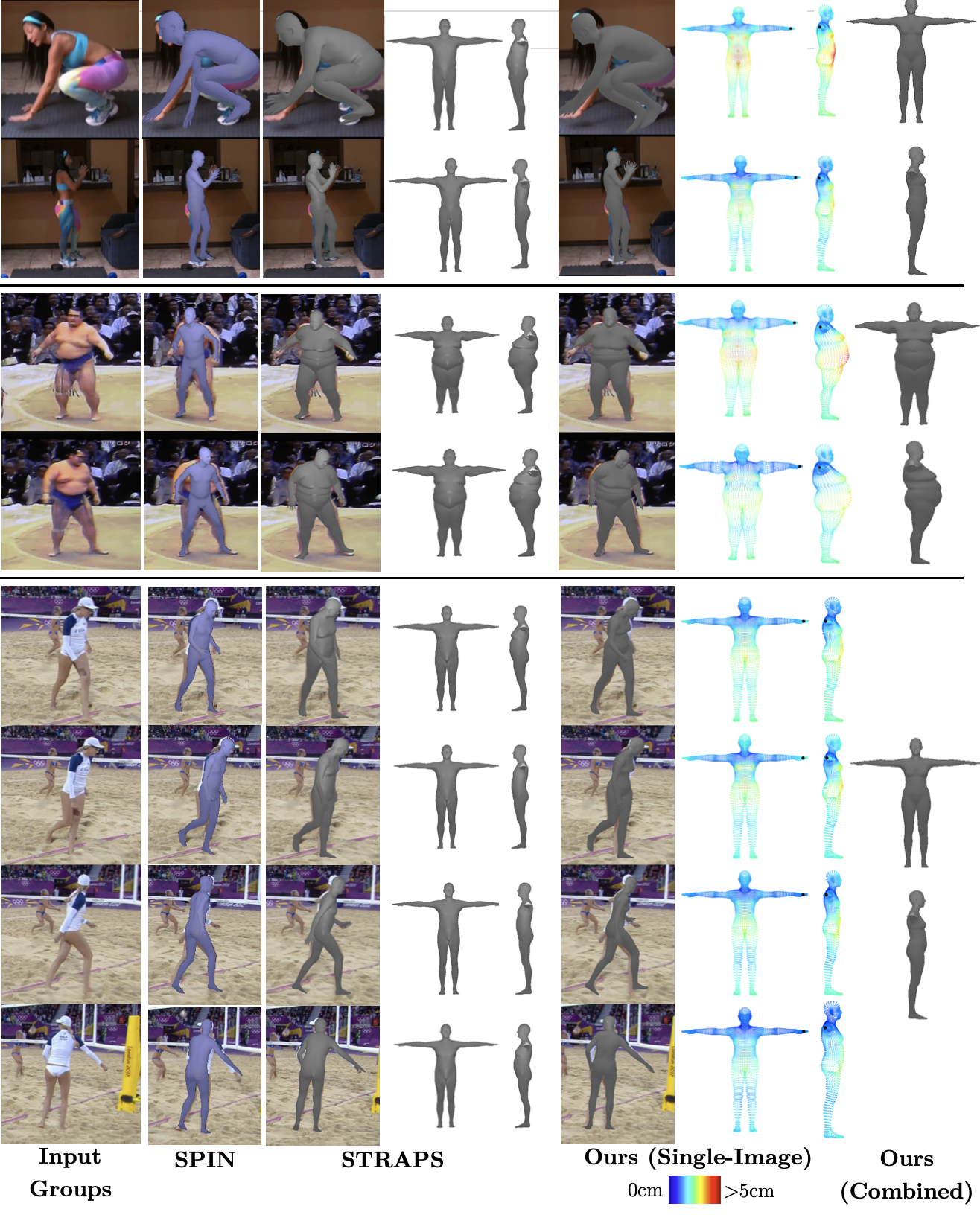}
    \caption{Example predictions on groups of images from SSP-3D \cite{STRAPS2020BMVC}. Single-image predictions from SPIN \cite{kolotouros2019spin}, STRAPS \cite{STRAPS2020BMVC} and our method are visualised, along with probabilistically combined body shapes from our method.}
    \label{fig:ssp3d_predictions}
    \vspace{-0.4cm}
\end{figure}

\noindent \textbf{Loss functions.} While inference occurs using a group of inputs, the model is trained with a dataset of \textit{individual} input-label pairs, $\{\mathbf{X}_k, \{\boldsymbol{\theta}_k, \boldsymbol{\beta}_k, \boldsymbol{\gamma}_k\}\}_{k=1}^K$, with $K$ i.i.d training samples. The negative log-likelihood is given by 
\begin{equation}
\begin{aligned}
\mathcal{L}_{\text{NLL}} &= - \sum_{k=1}^K \bigg( \log p(\boldsymbol{\theta}_k |\mathbf{X}_k) + \log p(\boldsymbol{\beta}_k |\mathbf{X}_k) \bigg)\\
& \propto \sum_{k=1}^K \bigg( \sum_{i=1}^{69} \log(2\pi \sigma^2_{\theta_i}) + \frac{(\theta_{k_i} - \mu_{\theta_i})^2}{\sigma^2_{\theta_i}} 
\\& \qquad \;\; + \sum_{j=1}^{10} \log(2\pi \sigma^2_{\beta_j}) + \frac{(\beta_{k_j} - \mu_{\beta_j})^2}{\sigma^2_{\beta_j}} \bigg)\\
\end{aligned}
\label{eqn:log_likelihood_loss}
\end{equation}
where $\mu_{\theta_i}$, $\mu_{\beta_i}$, $\sigma^2_{\theta_i}$ and $\sigma^2_{\beta_i}$ represent elements of the predicted SMPL mean and variance vectors $\boldsymbol{\mu}_\theta (\mathbf{X}_k, \mathbf{W})$, $\boldsymbol{\mu}_\beta (\mathbf{X}_k, \mathbf{W})$, $\boldsymbol{\sigma}^2_\theta (\mathbf{X}_k, \mathbf{W})$ and $\boldsymbol{\sigma}^2_\beta (\mathbf{X}_k, \mathbf{W})$, which are output by the neural network $f$ with weights $\mathbf{W}$. We maximise the log-likelihood of the model w.r.t $\mathbf{W}$ by minimising the loss function $\mathcal{L}_{\text{NLL}}$. Intuitively, each squared error term in Eqn. \ref{eqn:log_likelihood_loss} is adaptively-weighted by the corresponding predicted variance \cite{kendall2017whatuncertainties}. This mitigates the ill-posed nature of a naive squared error loss on SMPL parameters when training inputs are occluded, since the network learns to predict large variances for the parameters corresponding to invisible body parts, thus down-weighting the respective squared error terms. Furthermore, adaptive weighting means that our network is able to train stably without additional ``global'' losses on 3D vertices or 3D joints, as is common in most other recent methods \cite{STRAPS2020BMVC,pavlakos2018humanshape,kolotouros2019spin,hmrKanazawa17,zhang2019danet}.

Our network also predicts deterministic estimates of the global rotations $\boldsymbol{\gamma}_k$. Predictions $\boldsymbol{\hat{\gamma}}_k$ are supervised by
\begin{equation}
\mathcal{L}_{\text{glob}} = \sum_{k=1}^K \| \mathbf{R}(\boldsymbol{\gamma}_k) - \mathbf{R}( \boldsymbol{\hat{\gamma}}_k) \|_{F}^2.
\label{eqn:glob_loss}
\end{equation}
$\mathbf{R}(\boldsymbol{\gamma}) \in SO(3)$ is the rotation matrix corresponding to $\boldsymbol{\gamma}$.

Finally, our network estimates weak-perspective camera parameters $\mathbf{c}_k = [s_k, t^x_k, t^y_k]$, which are supervised using a 2D joint reprojection loss. Target 2D joint coordinates $\mathbf{J}_{k} \in \mathbb{R}^{L \times 2}$ are computed from $\{\boldsymbol{\theta}_k, \boldsymbol{\beta}_k,\boldsymbol{\gamma}_k\}$ during synthetic data generation (see Section \ref{subsec:network_training}). Predicted 2D joint coordinates are obtained by first differentiably sampling $\boldsymbol{\hat{\theta}}_k^i \sim p(\boldsymbol{\theta}_k |\mathbf{X}_k)$ and  $\boldsymbol{\hat{\beta}}_k^i \sim p(\boldsymbol{\beta}_k |\mathbf{X}_k)$ using the re-parameterisation trick \cite{kingma2014autoencoding}, for $i=1,...,B$ samples. These are converted into 2D joint samples using the SMPL model and weak-perspective projection
\begin{equation}
\mathbf{\hat{J}}_k^i = s_k \Pi(\mathcal{J}\mathcal{M}(\boldsymbol{\hat{\theta}}_k^i, \boldsymbol{\hat{\beta}}_k^i, \boldsymbol{\hat{\gamma}}_k)) + [t^x_k, t^y_k]
\end{equation}
where $\Pi()$ represents an orthographic projection. A squared error reprojection loss is imposed between the predicted 2D joint samples and the target 2D joints
\begin{equation}
\mathcal{L}_\text{2D} = \sum_{k=1}^{K} \sum_{i=1}^{B} \|\boldsymbol{\omega}_k(\mathbf{J}_k - \mathbf{\hat{J}}_k^i)\|_2^2
\label{eqn:2d_loss}
\end{equation}
where $\boldsymbol{\omega}_k \in \{0,1\}^L$ denote the visibilities of the target joints (1 if visible, 0 otherwise), which are computed during synthetic data generation. We apply a reprojection loss on \textit{samples} from the predicted body shape and pose distributions, instead of only on the means of the distributions, because \textit{any} 3D body sampled from the distributions must match the 2D joint locations present in the input $\mathbf{X}_k$.

Our overall loss function is given by $\mathcal{L} = \mathcal{L}_\text{NLL} + \lambda_\text{glob} \mathcal{L}_\text{glob} + \lambda_\text{2D} \mathcal{L}_\text{2D}$ where $\lambda_\text{glob}, \lambda_\text{2D}$ are weighting terms.

\noindent \textbf{Synthetic data generation.} To train our distribution prediction network using the proposed losses, we require training data consisting of input proxy representations paired with target SMPL shape, pose and global rotation parameters, $\{\mathbf{X}_k, \{\boldsymbol{\theta}_k, \boldsymbol{\beta}_k, \boldsymbol{\gamma}_k\}\}_{k=1}^K$. We employ a similar synthetic training data generation process as \cite{STRAPS2020BMVC}. In short, within each iteration of the training loop, $\boldsymbol{\theta}_k$ and $\boldsymbol{\gamma}_k$ are sampled from any suitable dataset with SMPL pose parameters \cite{h36m_pami,Lassner:UP:2017,AMASS:2019,vonMarcard2018}, while $\boldsymbol{\beta}_k$ are randomly sampled from a suitably high-variance Gaussian distribution to ensure body shape diversity. These are converted into synthetic silhouette and joint heatmap representations $\mathbf{X}_k$ and target 2D joint coordinates $\mathbf{J}_{k}$ using the SMPL model, a renderer \cite{kato2018renderer} and randomly sampled perspective camera parameters. The clean synthetic inputs are corrupted to model the failure modes of the off-the-shelf detection and segmentation CNNs used at test-time, such as noisy keypoint locations, and occluded silhouettes. Examples are given in Figure \ref{fig:synthtic_data_example_predictions}. 

We improve the data generation process of \cite{STRAPS2020BMVC} in two ways. First, we significantly increase the severity of the occlusion and cropping augmentations to match the occlusions seen in challenging test datasets such as 3DPW (illustrated in Figure \ref{fig:3dpw_uncertainty}, first row). Second, we explicitly compute a joint visibility vector $\boldsymbol{\omega}_k$ (1 if visible, 0 otherwise) for each $\mathbf{J}_k$ and set the heatmaps corresponding to invisible joints to 0, unlike \cite{STRAPS2020BMVC}. This is necessary for our distribution prediction network to learn to be uncertain about pose parameters corresponding to invisible body parts.

\begin{table}[t]
\centering
\small
\renewcommand{\tabcolsep}{1pt}
\begin{tabular}{l l c c} 
 \hline
 \textbf{Input groups} & \textbf{Method} & \textbf{Synthetic} & \textbf{Synthetic}\\
 & & \textbf{Clean} & \textbf{Corrupted}\\
 & & PVE-T-SC & PVE-T-SC\\ [0.5ex] 
 \hline
 \textbf{Single-Input} & Ours & 14.4 & 15.1\\
 \hline
 \multirow{2}{0.3\linewidth}{\textbf{Quadruplets} Front + L/R Side + Back}\\ 
 & Ours + Mean & 13.1 & 13.3\\
 & Ours + PC & \textbf{13.0} & \textbf{12.8}\\
 \hline
 \textbf{Pairs} \\
 Front + L Side or  & Ours + PC & 13.5 & 13.4\\
 Back + R Side & & &\\
 \hline
 \textbf{Pairs} \\
 Front + Back or & Ours + PC & 13.6 & 13.8\\
 L + R Side & & &\\
 \hline
\end{tabular}
\caption{PVE-T-SC (mm) results on synthetic data (see Figure \ref{fig:synthtic_data_example_predictions})  investigating: i) probabilistic shape combination (PC) versus simple averaging (Mean), ii) effect of increasing input group size from 1 to 2 (Pairs) to 4 (Quadruplets) and iii) effect of global rotation variation within pairs of inputs.}
\label{table:synth_ablation_comparison}
\end{table}


\section{Implementation Details}
\noindent \textbf{Network Architecture.} We use a ResNet-18 \cite{He2015} encoder followed by a multi-layer perceptron (MLP) to predict SMPL parameter distributions. The MLP is comprised of one hidden layer with 512 neurons and ELU \cite{clevert2016elu} activation and one output layer with 164 neurons, which predicts the set of outputs $\mathbf{Y}$. Predicted variances are forced to be positive using an exponential activation function.

\noindent \textbf{Training dataset.} Synthetic training data is generated by sampling SMPL pose parameters from the training sets of UP-3D \cite{Lassner:UP:2017}, 3DPW \cite{vonMarcard2018}, and Human3.6M \cite{h36m_pami} (Subjects 1, 5, 6, 7, 8).

\noindent \textbf{Training details.} We use Adam \cite{kingma2014adam} with a learning rate of 1e-4 and a batch size of 120, and train for 100 epochs, which takes 1.5 days on a 2080Ti GPU. Inference runs at ~4fps, 90\% of which is silhouette and joint prediction \cite{Guler2018DensePose, wu2019detectron2}.

\begin{table}[t!]
\centering
\small
\begin{tabular}{c l c} 
 \hline
 \multirow{2}{0.23\linewidth}{\textbf{Max. input group size}} & \textbf{Method} & \textbf{SSP-3D}\\
 & & PVE-T-SC\\ [0.5ex] 
 \hline
 1 & Ours & 15.2\\
 2 & Ours + PC & 13.9\\
 3 & Ours + PC & 13.6\\
 4 & Ours + PC & 13.5\\
 \hline
 5 & Ours + Mean & 13.6\\
 & Ours + PC & \textbf{13.3}\\
 \hline
\end{tabular}
\caption{PVE-T-SC (mm) results on SSP-3D \cite{STRAPS2020BMVC} comparing i) probabilistic shape combination (PC) versus simple averaging (Mean) and ii) effect of increasing input group size from 1 to 5.}
\vspace{-0.3cm}
\label{table:ssp3d_ablation_comparison}
\end{table}

\noindent \textbf{Evaluation datasets.} We use the test set of 3DPW to evaluate pose prediction accuracy. It consists of 35515 RGB images of 7 subjects with paired ground-truth SMPL parameters. We report mean per joint position error after scale correction (MPJPE-SC) and after Procrustes analysis (MPJPE-PA). We use the scale correction technique introduced in \cite{STRAPS2020BMVC} to combat the ambiguity between subject scale and distance from camera. MPJPE-SC measures 3D joint error up to scale and MPJPE-PA measures 3D joint error up to scale and global rotation. We also report scale-corrected per-vertex Euclidean error in a T-pose (PVE-T-SC) on the SSP-3D dataset \cite{STRAPS2020BMVC} to evaluate identity-dependent shape prediction accuracy. SSP-3D consists of 311 images of 62 subjects and pseudo-ground-truth SMPL parameters.

In addition, we evaluate body shape prediction accuracy on a private dataset consisting of 6 subjects (4 male, 2 female) with 4 RGB images of each and ground-truth body measurements obtained using a tape measure or full 3D body scanning technology. The subjects' body poses, clothing, surrounding environments and camera viewpoints vary between images. Example images are in the supplementary material.

Finally, we create a synthetic dataset for our experimental analysis. It consists of 1000 synthetic humans with randomly sampled SMPL body shapes, each posed in 4 different SMPL poses sampled from Human3.6M \cite{h36m_pami} subjects 9 and 11. Global orientations are set such that the camera is facing the human's front, back, left or right. A group of 4 clean synthetic inputs and 4 corrupted inputs is rendered for each human, where the corruptions used are the same as the data augmentations applied during training. Examples are given in Figure \ref{fig:synthtic_data_example_predictions}.

\begin{table}[t]
\centering
\small
\begin{tabular}{c l c c} 
 \hline
 \multirow{2}{0.23\linewidth}{\textbf{Max. input group size}} & \textbf{Method} & \textbf{SSP-3D}\\
 & & PVE-T-SC\\ [0.5ex] 
 \hline
 & HMR \cite{hmrKanazawa17} & 22.9\\
 & GraphCMR \cite{kolotouros2019cmr} & 19.5\\
  \large 1 & SPIN \cite{kolotouros2019spin} & 22.2\\
 & DaNet \cite{zhang2019danet} & 22.1\\
 & STRAPS \cite{STRAPS2020BMVC} & 15.9\\
 & Ours & \textbf{15.2}\\
 \hline
 & HMR \cite{hmrKanazawa17} + Mean & 22.9\\
 & GraphCMR \cite{kolotouros2019cmr} + Mean & 19.3\\
 & SPIN \cite{kolotouros2019spin} + Mean & 21.9\\
 \large 5 & DaNet \cite{zhang2019danet} + Mean & 22.1\\
 & STRAPS \cite{STRAPS2020BMVC} + Mean & 14.4 \\
 & Ours + Mean & 13.6\\
 & Ours + PC & \textbf{13.3}\\
 \hline
 Video & VIBE \cite{kocabas2019vibe} & 24.1\\
 \hline
\end{tabular}
\caption{Comparison with the state-of-the-art in terms of PVE-T-SC (mm) on SSP-3D \cite{STRAPS2020BMVC}. Our method surpasses the state-of-the-art when using single-image inputs. Probabilistic shape combination (PC) outperforms simple averaging of predictions from other methods when using groups of up to 5 images, as well as video predictions from \cite{kocabas2019vibe}.}
\label{table:ssp3d_sota_comparison}
\end{table}

\begin{table}[t]
\centering
\small
\begin{tabular}{l c c c} 
 \hline
 \textbf{Method} & \multicolumn{2}{c}{\textbf{3DPW}}\\
 & MPJPE-SC & MPJPE-PA\\ [0.5ex] 
 \hline
 HMR \cite{hmrKanazawa17} & 102.8 & 71.5\\
 GraphCMR \cite{kolotouros2019cmr} & 102.0 & 70.2\\
 SPIN \cite{kolotouros2019spin} & 89.4 & 59.2\\
 I2L-MeshNet \cite{Moon_2020_ECCV_I2L-MeshNet} & - & 57.7\\
 Biggs et al. \cite{biggs2020multibodies} & - & 55.6\\
 DaNet \cite{zhang2019danet} & \textbf{82.4} & \textbf{54.8}\\
 \hline
 HMR (unpaired) \cite{hmrKanazawa17} & 126.3 & 92.0\\
 Kundu et al. \cite{kundu_human_mesh} & - & 89.8\\
 STRAPS \cite{STRAPS2020BMVC} & 99.0 & 66.8\\
 Ours & 90.9 & 61.0\\
 \hline
\end{tabular}
\caption{Comparison with the state-of-the-art in terms of MPJPE-SC and MPJPE-PA (both mm) on 3DPW \cite{vonMarcard2018}. Methods in the top half require training images paired with 3D ground-truth, methods in the bottom half do not.}
\label{table:3dpw_sota_comparison}
\vspace{-0.3cm}
\end{table}

\begin{figure}[t]
    \centering
    \includegraphics[width=\linewidth]{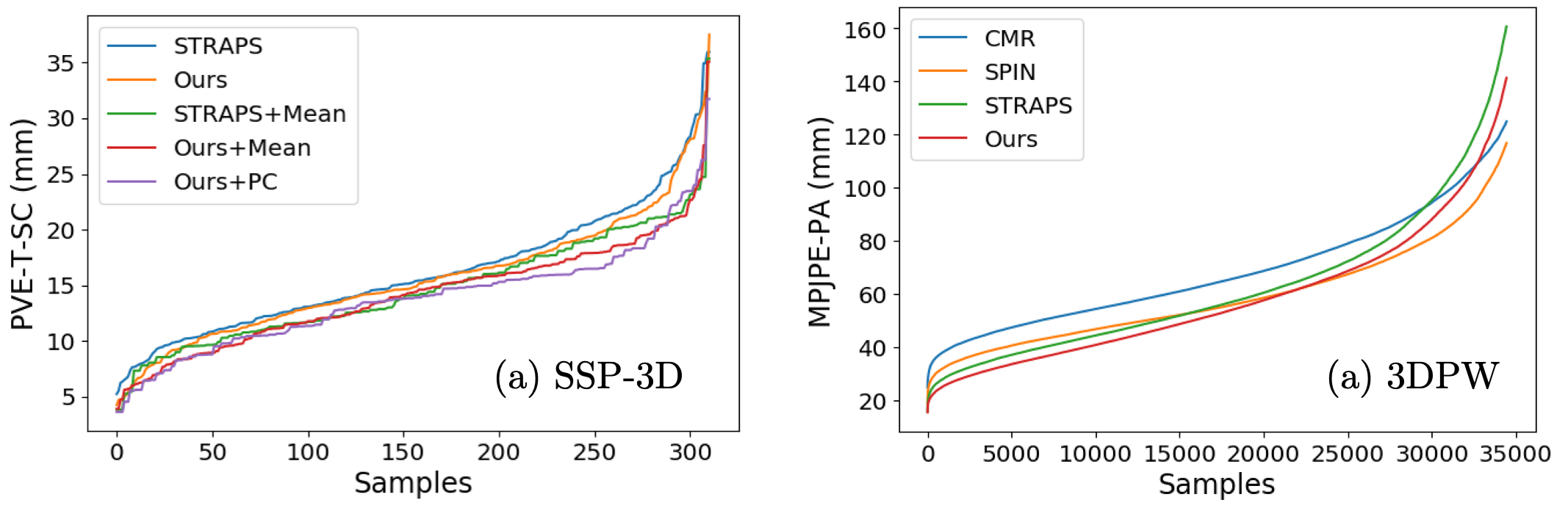}
    \caption{Comparison with other methods using sorted distributions of a) PVE-T-SC per SSP-3D evaluation sample and b) MPJPE-PA per 3DPW evaluation sample.}
    \label{fig:3dpw_ssp3d_sorted_error_distribution_combined}
    \vspace{-0cm}
\end{figure}

\section{Experimental Results}
\label{sec:experimental_results}

In this section, we present our ablation studies, where we investigate uncertainty predictions, compare probabilistic shape combination with simple averaging and explore the effects of varying input group sizes and global rotation variation within groups. We also compare our method to other approaches in terms of shape and pose accuracy.

\subsection{Ablation studies}

\noindent \textbf{Pose and shape uncertainty.} SMPL pose and shape prediction uncertainties are represented by the predicted variances $\boldsymbol{\sigma}^2_\theta$ and $\boldsymbol{\sigma}^2_\beta$. Rows 2 and 5 of Figure \ref{fig:synthtic_data_example_predictions} illustrate pose prediction uncertainty on clean and corrupted synthetic inputs. Heavily-occluded inputs (e.g. the corrupted input in column 4), result in large predicted variances for the pose parameters corresponding to the occluded joints, while predicted variances are smaller for visible joints. This behaviour is replicated on real inputs from the 3DPW dataset, as shown by Figure \ref{fig:3dpw_uncertainty} where the network is uncertain about the 3D locations of occluded and out-of-frame body parts.

Figure \ref{fig:synthtic_data_example_predictions} also showcases shape parameter prediction uncertainty on synthetic data, in rows 3 and 6. The network is more uncertain about body shape when the subject is heavily occluded and/or in a challenging pose, seen by comparing the sitting pose in column 2 with the standing poses in columns 1 and 3. This behaviour is also seen on real inputs in Figure \ref{fig:ssp3d_predictions}, e.g. by comparing the crouching pose in row 1 with the standing pose in row 2.

\noindent \textbf{Body shape combination method.} We compare probabilistic body shape combination (from Section \ref{subsec:body_shape_combination}) with a simpler heuristic combination, where we obtain combined body shape estimates from a group of inputs $\{\mathbf{X}_n\}_{n=1}^N$ by simply averaging (i.e. taking the mean of) the shape distribution means $\{\boldsymbol{\mu}_\beta(\mathbf{X}_n)\}_{n=1}^N$. Rows 3-4 in Table \ref{table:synth_ablation_comparison} show that better shape estimation metrics are attained using probabilistic combination versus simple averaging on synthetic input quadruplets (examples in Figure \ref{fig:synthtic_data_example_predictions}). This is replicated on groups of real inputs from SSP-3D, as shown in Table \ref{table:ssp3d_ablation_comparison}, row 5 versus row 6. Since probabilistic combination may be interpreted as uncertainty-weighted averaging (Eqn. \ref{eqn:combined_shape_mean_var}), these experiments suggest that inaccurate mean body shape predictions are generally accompanied by large prediction uncertainty, and subsequently down-weighted during probabilistic combination. This may explain why probabilistic combination actually gives better shape metrics when evaluating on corrupted synthetic inputs compared to clean inputs in Table \ref{table:synth_ablation_comparison}, since heavy input corruption results in inaccurate but highly-uncertain shape estimates.

\noindent \textbf{Input group size.} Table \ref{table:synth_ablation_comparison} also investigates the effect of the input group size, evaluated on our synthetic dataset, by comparing single inputs (i.e. group size of 1) with body shape combination applied to pairs and quadruplets (i.e. input group sizes of 2 and 4). Body shape metrics are significantly improved when using pairs compared to single images, suggesting that probabilistic combination is successfully using shape information from the multiple inputs. A smaller improvement is seen when using quadruplets versus pairs. Table \ref{table:ssp3d_ablation_comparison} shows that increasing the input group size on real data (from SSP-3D) also results in a consistent but diminishing improvement in shape prediction metrics.

\noindent \textbf{Global rotation variation.} To investigate whether variation in global rotation of the subject between inputs is correlated with shape prediction accuracy, we split each group of 4 inputs from our synthetic dataset into groups of 2 in two ways: (front, left) + (back, right) and (front, back) + (left, right). We expect the latter split to be less informative for shape prediction as the pairs contain more redundant visual shape information. This is corroborated by the experiments labelled ``Pairs'' in Table \ref{table:synth_ablation_comparison}, where the former split yields better shape metrics, particularly for corrupted inputs where the amount of visual shape information in each individual input is lower.

\begin{table}[t!]
\centering
\small
\renewcommand{\tabcolsep}{3.5pt}
\begin{tabular}{c l c c c c c c} 
 \hline
 \multirow{2}{0.12\linewidth}{\textbf{Group size}} & \textbf{Method} & \multicolumn{6}{c}{\textbf{RMSE}}\\
 & & C & S & H & B & F & T \\
 \hline
\multirow{3}{3em}{1} & SPIN \cite{kolotouros2019spin} & 6.9 & 8.0 & 6.6 & 6.9 & 2.5 & 5.3\\
 & STRAPS \cite{STRAPS2020BMVC} & 6.7 & 5.3 & 4.3 & 3.9 & 1.8 & 3.7\\
 & Ours & 4.9 & 4.7 & 5.5 & 4.2 & 1.8 & 3.9\\
 \hline
\multirow{3}{3em}{4} & SPIN \cite{kolotouros2019spin} + Mean & 6.5 & 8.1 & 6.4 & 6.7 & 2.4 & 5.1\\
 & STRAPS \cite{STRAPS2020BMVC} + Mean & 6.1 & 4.2 & 4.0 & \textbf{3.2} & 1.7 & 3.3\\
 & Ours + Mean & 3.4 & 3.9 & 3.8 & 4.9 & \textbf{1.6} & 3.1\\
 & Ours + PC & \textbf{3.1} & \textbf{3.8} & \textbf{2.7} & 5.0 & 1.7 & \textbf{2.8}\\
 \hline
\end{tabular}
\caption{Comparison with the state-of-the-art in terms of tape measurement RMSE (cm) on our private shape evaluation dataset. Errors are reported for the chest (C), stomach (S), hips (H), biceps (B), forearms (F) and thighs (T).}
\label{table:size_stream_results}
\end{table}


\subsection{Comparison with the state-of-the-art}

\noindent \textbf{Shape prediction.} Our method surpasses the state-of-the-art on SSP-3D in the single-input case (group size of 1), as shown in Table \ref{table:ssp3d_sota_comparison} and Figure \ref{fig:ssp3d_predictions}. The improvement over the similar synthetic training method in STRAPS \cite{STRAPS2020BMVC} is primarily due to our improved training data augmentations. When using groups of multiple images as inputs (with group size = 5), probabilistic combination outperforms simple averaging of predictions from all other methods. The distribution of errors per SSP-3D sample, shown in Figure \ref{fig:3dpw_ssp3d_sorted_error_distribution_combined}, suggests that probabilistic combination particularly improves errors for challenging samples, where the uncertainty weighting is more meaningful.

Table \ref{table:size_stream_results} compares tape measurement errors computed using shape predictions from our method and competitors on a private dataset. Probabilistic combination results in the lowest measurement errors for large body parts, such as the chest, stomach, waist and hips. However, it is less accurate on smaller body parts (e.g. biceps and forearms), which are significantly obscured by clothing. In general, our method may over-estimate measurements for subjects with loose clothing, since silhouette-based inputs don't distinguish between clothing and the human body.

\noindent \textbf{Pose prediction.} While we focus on body shape estimation, Table \ref{table:3dpw_sota_comparison} shows that our method is competitive with the state-of-the-art on 3DPW, and surpasses other methods that do not require training images paired with 3D labels. Figure \ref{fig:3dpw_ssp3d_sorted_error_distribution_combined} demonstrates that our method does well on low-to-medium difficulty samples, but struggles with the most challenging ones, which typically exhibit very severe occlusions leading to degraded proxy representations. Nevertheless, we outperform STRAPS \cite{STRAPS2020BMVC} on these challenging samples due to improved data augmentation and the adaptive loss weighting discussed in Section \ref{subsec:network_training}, which results in a more stable improvement of pose metrics during training.

\section{Conclusion}
In this paper, we have proposed the novel task of human body shape estimation from a group of images of the same subject, without imposing any constraints on body pose, camera viewpoint or backgrounds. Our solution predicts multivariate Gaussian distributions over SMPL \cite{SMPL:2015} body shape and pose parameters conditioned on the input images. We probabilistically combine predicted body shape distributions from each image to obtain a final multi-image shape prediction, and experimentally show that probabilistically combined estimates are more accurate than both individual single-image predictions, as well as naively-averaged single-image predictions, when evaluated on SSP-3D and a private dataset of tape-measured humans. Furthermore, predicting distributions over SMPL parameters allows us to estimate the heteroscedastic aleatoric uncertainty associated with pose predictions, which is useful when faced with input images containing occluded or out-of-frame body parts. Future work can consider using (i) clothed and textured synthetic data to further close the synthetic-to-real domain gap, and (ii) more expressive predicted distributions than the simple Gaussians proposed in this paper.

\noindent \textbf{Acknowledgements.} We thank Dr. Yu Chen (Metail) and Dr. David Bruner (SizeStream) for providing body shape evaluation data and 3D body scans. This research was sponsored by SizeStream UK Ltd.

\clearpage

\twocolumn[
\large
\centering
\textbf{Supplementary Material: Probabilistic 3D Human Shape and Pose Estimation from Multiple Unconstrained Images in the Wild\\}
\vspace{0.2in}
]
\noindent This document provides additional material supplementing the main manuscript. Section \ref{sec:supmat_implementation_details} contains details regarding training data generation, evaluation protocols and probabilistic shape combination. Section \ref{sec:supmat_experiment_results} discusses qualitative results on the SSP-3D \cite{STRAPS2020BMVC} and 3DPW \cite{vonMarcard2018} datasets, as well as providing examples from our private evaluation dataset of tape-measured humans.

\section{Implementation Details}
\label{sec:supmat_implementation_details}

\noindent \textbf{Training.} Table \ref{table:sup_mat_augment_hypparams} lists the data augmentation methods used to bridge the synthetic-to-real domain gap during synthetic training data generation, along with associated hyperparameter values. Table \ref{table:sup_mat_hypparams} lists additional hyperparameter values not given in the main manuscript.

\noindent \textbf{Uncertainty Visualisation.} Figures \ref{fig:sup_mat_ssp3d} and \ref{fig:sup_mat_3dpw} in this supplementary material, as well as several figures in the main manuscript, visualise per-vertex prediction uncertainties. These are computed from the predicted SMPL \cite{SMPL:2015} pose and shape parameter distributions by i) sampling 100 SMPL parameter vectors from the predicted distributions, ii) passing each of these samples through the SMPL function to get the corresponding vertex meshes, iii) computing the mean location of each vertex over all the samples and iv) determining the average Euclidean distance from the mean for each vertex over all the samples, which is ultimately visualised in the scatter plots as a measure of uncertainty.

\noindent \textbf{SSP-3D Evaluation Groups.} SSP-3D \cite{STRAPS2020BMVC} contains 311 images of 62 subjects, where subjects can have a different number of associated images. To evaluate our multi-input shape prediction method, the images for each subject were split into groups of \textit{maximum} size equal to $N$, where $N$ ranged from 1 to 5. For example, if a subject has 6 associated images and $N=4$, the images would be split into two groups with 4 and 2 images respectively. Splitting/group assignment was done after random shuffling of the images to prevent sequential images with similar poses/global orientations from always being in the same group.

\begin{figure}
    \centering
    \includegraphics[width=\linewidth]{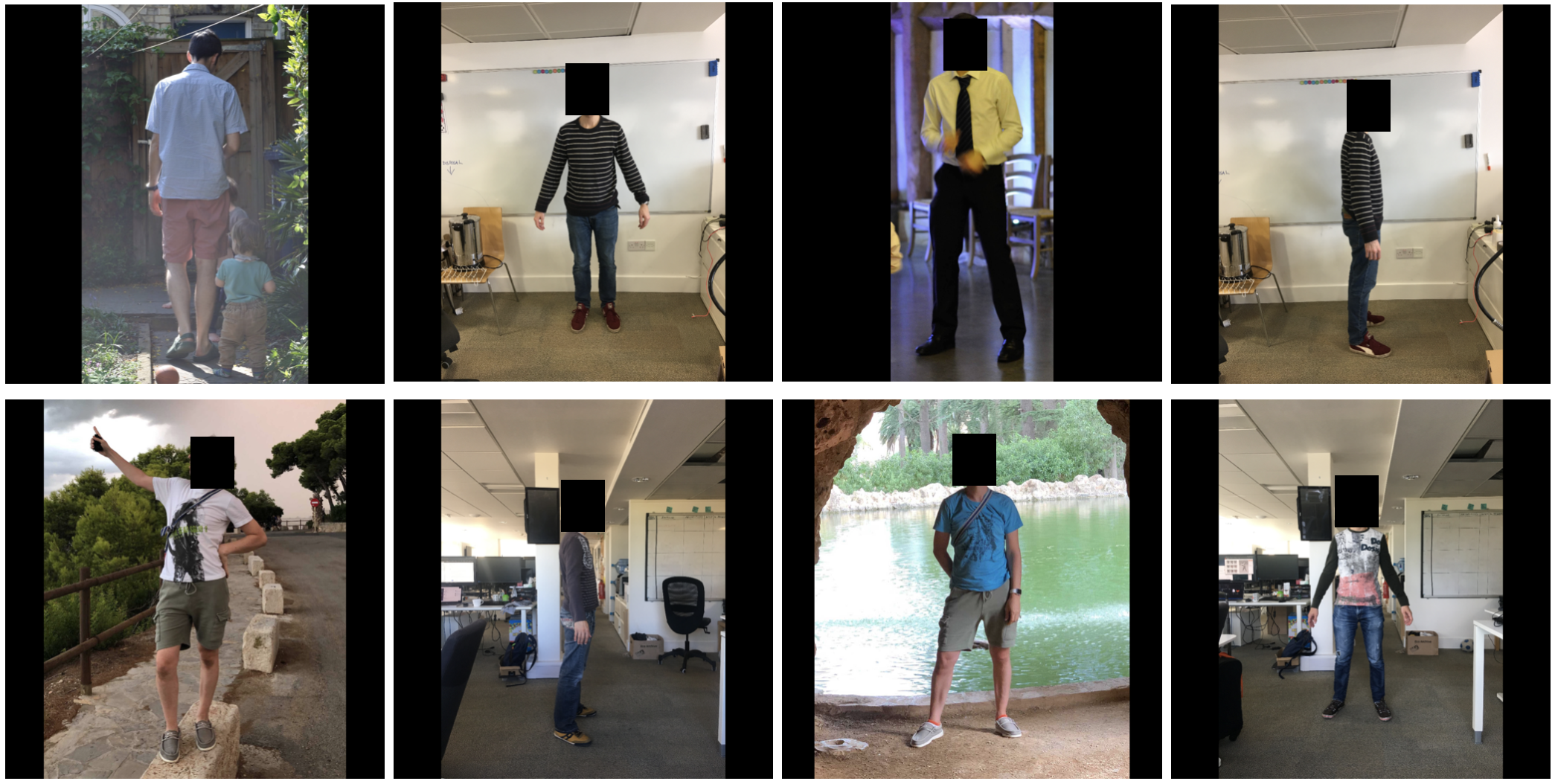}
    \caption{Example images from our private dataset of humans with body measurements obtained using a tape measure or 3D body scanners. The subjects' body pose, clothing, surrounding environment and camera viewpoints vary between images.}
    \label{fig:supmat_sizestream}
\end{figure}

\noindent \textbf{Tape measurement normalisation by height.} There is an inherent ambiguity between 3D subject size/scale and distance from camera. Since the true camera location relative to the 3D subject (and the focal length) is unknown, it is not possible to estimate the absolute size of the subject given an image. This is accounted for by the PVE-T-SC \cite{STRAPS2020BMVC} metric used to evaluate shape prediction accuracy on synthetic data and SSP-3D in the main manuscript. For our evaluation dataset of tape-measured humans (see Figure \ref{fig:supmat_sizestream}), scale correction is done using the subject's height. The height of the predicted SMPL human can be determined by computing the neutral-pose mesh (i.e. pose parameters/joint rotations set to 0) and measuring the $y$-axis distance between the top of the head and bottom of the feet. The ratio between the subject's true height and this predicted height is then used to scale all the predicted body measurements derived from the neutral-pose mesh.

\begin{table}[t]
\centering
\small
\begin{tabular}{l l c}
    \hline
    \noalign{\smallskip} 
    \textbf{Augmentation} & \textbf{Hyperparameter} & \textbf{Value}\\
    \noalign{\smallskip}
    \hline
    \noalign{\smallskip}
    Body part occlusion & Occlusion prob. & 0.1 \\
    2D joints L/R swap & Swap prob. & 0.1\\
    Half-image occlusion & Occlusion prob. & 0.05\\
    2D joints removal & Removal prob. & 0.1\\
    2D joints noise & Noise range & [-8, 8] pixels\\
    2D vertices noise & Noise range & [-10, 10] mm \\
    Occlusion box & Probability, Size & 0.5, 48 pixels \\
    \noalign{\smallskip}
    \hline
    \noalign{\smallskip}
    \noalign{\smallskip}
    \end{tabular}
\caption{List of synthetic training data augmentations and their associated hyperparameter values. Body part occlusion uses the 24 DensePose \cite{Guler2018DensePose} parts. Joint L/R swap is done for shoulders, elbows, wrists, hips, knees, ankles.}
\label{table:sup_mat_augment_hypparams}
\end{table}

\begin{table}[t]
\centering
\small
\begin{tabular}{l c}
    \hline
    \noalign{\smallskip} 
    \textbf{Hyperparameter} & \textbf{Value}\\
    \noalign{\smallskip}
    \hline
    \noalign{\smallskip}
    Shape parameter sampling mean & 0 \\
    Shape parameter sampling var. & 2.25 \\
    Cam. translation sampling mean & (0, -0.2, 2.5) m\\
    Cam. translation sampling var. & (0.05, 0.05, 0.25) m\\
    Cam. focal length & 300.0\\
    Proxy representation dimensions & $256 \times 256$ pixels\\
    2D joint confidence threshold & 0.025\\
    \noalign{\smallskip}
    \hline
    \noalign{\smallskip}
    \noalign{\smallskip}
    \end{tabular}
\caption{List of hyperparameter values not provided in the main manuscript.}
\label{table:sup_mat_hypparams}
\end{table}

\noindent \textbf{Probabilistic shape combination.} The main manuscript presents our method to probabilistically combine individual body shape distributions, $p(\boldsymbol{\beta} | \mathbf{X}_n)$ for $n = 1, ..., N$, into a final distribution $p(\boldsymbol{\beta} | \{\mathbf{X}_n\}_{n=1}^N)$. The full derivation is given below:
\begin{equation}
\begin{aligned}
p(\boldsymbol{\beta} | \{\mathbf{X}_n\}_{n=1}^N) 
& \propto p(\{\mathbf{X}_n\}_{n=1}^N | \boldsymbol{\beta})p(\boldsymbol{\beta})\\
& = \bigg(\prod_{n=1}^N p(\mathbf{X}_n | \boldsymbol{\beta}) \bigg)p(\boldsymbol{\beta})\\
& \propto \frac{\prod_{n=1}^N p(\boldsymbol{\beta} | \mathbf{X}_n)}{p(\boldsymbol{\beta})^{N-1}}\\
& \propto \prod_{n=1}^N p(\boldsymbol{\beta} | \mathbf{X}_n).
\end{aligned}
\label{eqn:probabilistic_shape_combination}
\vspace{-0.1cm}
\end{equation}
The first and third lines use Bayes' theorem. The second line follows from the conditional independence assumption $(\mathbf{X}_i \indep \mathbf{X}_j) | \boldsymbol{\beta}$ for $i,j \in \{1,...,N\}$ and $i \neq j$. This assumption is reasonable because only the subject's body shape is fixed across inputs - hence, the inputs are independent given the body shape parameters. The final line follows from assuming an (improper) uniform prior over the shape parameters $p(\boldsymbol{\beta}) = 1$.

\section{Experimental Results}
\label{sec:supmat_experiment_results}

\begin{table*}[!b]
\centering
    \begin{tabular}{l cc cc}
    \hline
    \multirow{2}{3em}{\textbf{Input}} & \multicolumn{2}{c}{\textbf{3DPW}} & \multicolumn{2}{c}{\textbf{SSP-3D}}\\
    & MPJPE-SC & MPJPE-PA & PVE-PA & PVE-T-SC\\
    \noalign{\smallskip}
    \hline
    \noalign{\smallskip}
    \textbf{GT Synthetic} Silh. + 2D Joint Heatmaps & \textbf{64.3} & \textbf{45.7} & \textbf{52.9} & \textbf{10.1}\\
    \noalign{\smallskip}
    \textbf{GT} Silh. + 2D Joint Heatmaps & - & - & 69.9 & 14.4\\
    \noalign{\smallskip}
    \textbf{Predicted} Silh. + 2D Joint Heatmaps & 90.9 & 61.0 & 71.4 & 15.2\\
    \noalign{\smallskip}
    \hline
    \end{tabular}
\caption{Comparison between ground-truth (GT), synthetic ground-truth and predicted input silhouettes and 2D joints, in terms of MPJPE-SC and MPJPE-PA (both in mm) on 3DPW \cite{vonMarcard2018}, as well as PVE-PA and PVE-T-SC (both in mm) on SSP-3D \cite{STRAPS2020BMVC}. Predicted silhouettes are obtained using DensePose \cite{Guler2018DensePose} and predicted 2D joint coordinates and confidences (for thresholding) are obtained using Keypoint-RCNN from Detectron2 \cite{wu2019detectron2}. Synthetic ground-truth inputs are obtained by rendering the SMPL \cite{SMPL:2015} body mesh labels given by SSP-3D and 3DPW, using ground-truth camera parameters, into silhouette and 2D joint input representations.}
\label{table:supmat_gt_vs_pred_inputs}
\end{table*}

\begin{figure*}
    \centering
    \includegraphics[width=\linewidth]{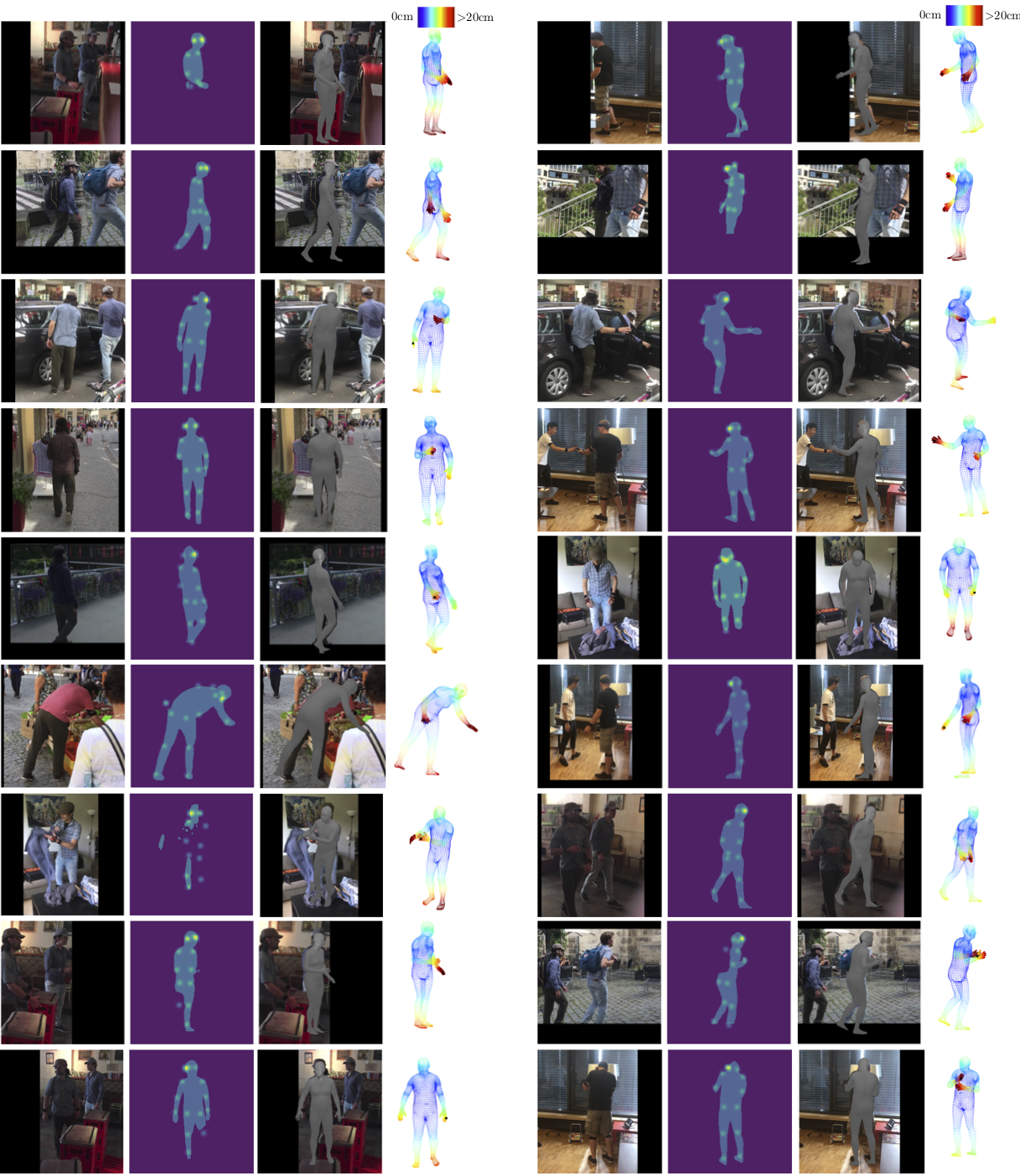}
    \caption{Qualitative results using single-image inputs from 3DPW \cite{vonMarcard2018}. Black dots indicate left hands. On each side: 1st column is input image, 2nd column is predicted proxy representation, 3rd column is predicted SMPL body and 4th column is 3D per-vertex uncertainty in the SMPL pose and shape distribution prediction. Vertices of occluded and out-of-frame body parts have higher prediction uncertainties.}
    \label{fig:sup_mat_3dpw}
\end{figure*}

\noindent \textbf{Evaluation using ground-truth vs predicted inputs.} The synthetic training data augmentations listed in Table \ref{table:sup_mat_augment_hypparams} and the main manuscript are used to increase the robustness of our distribution prediction neural network to noisy and occluded test data, as demonstrated in Figure \ref{fig:sup_mat_3dpw}. However, the synthetic-to-real domain gap still persists, as evidenced by Table \ref{table:supmat_gt_vs_pred_inputs}, which compares body shape and pose prediction metrics when using ground-truth, synthetic ground-truth and predicted input proxy representations. A significant improvement in both body shape and pose metrics is observed when using synthetic inputs, instead of predicted inputs. This is mostly because predicted input silhouettes and 2D joints can be very inaccurate in cases with challenging poses, significant occlusion or occluding humans, such that the synthetic training data augmentations are not sufficient.  Moreover, synthetic SMPL human silhouettes are not clothed, while silhouette predictors generally classify clothing pixels as part of the human body. This is particularly detrimental to body shape prediction metrics when subjects are dressed in loose clothing, as can be seen in Figure \ref{fig:sup_mat_3dpw} (left side, rows 3 and 4), where our method tends to over-estimate the subject's body proportions.

\begin{figure*}
    \centering
    \includegraphics[width=0.9\linewidth]{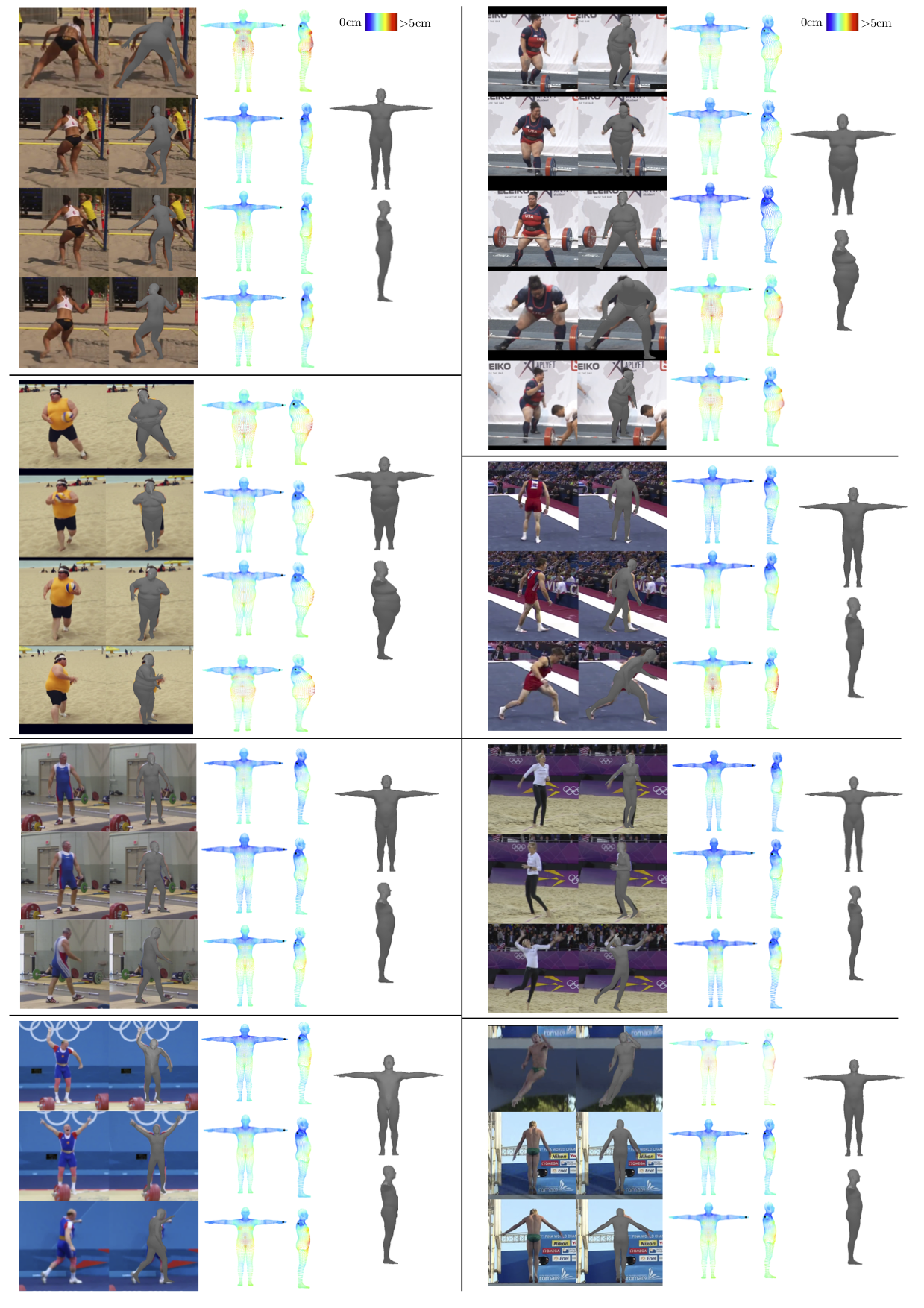}
    \vspace{-0.25in}
    \caption{Qualitative results on groups of input images from SSP-3D \cite{STRAPS2020BMVC}. Black dots indicate left hands. Within each cell: 1st column is group of input images, 2nd column is predicted SMPL body, 3rd and 4th columns show 3D per-vertex uncertainty in the SMPL \textit{shape} distribution prediction, 5th column is the probabilistically-combined body shape. Challenging poses lead to large shape prediction uncertainty.}
    \label{fig:sup_mat_ssp3d}
\end{figure*}

\noindent \textbf{SSP-3D qualitative results.}
Figure \ref{fig:sup_mat_ssp3d} shows qualitative results, particularly focusing on shape prediction, on groups of input images from SSP-3D \cite{STRAPS2020BMVC} corresponding to subjects with a wide range of body shapes. The first column in each cell shows the input images in the group. The second column shows the predicted SMPL \cite{SMPL:2015} body (rendered) for each \textit{individual} image, obtained by passing the mean of predicted SMPL parameter distributions through the SMPL function. The third and fourth columns visualise the 3D per-vertex uncertainty (or variance) in the individual SMPL shape distribution predictions (in a neutral pose i.e. pose parameters/joint rotations set to 0). The fifth column shows the \textit{combined} body shape prediction, which are obtained by probabilistically combining the individual shape distributions. 

In particular, note the relationship between challenging poses with significant self-occlusion (e.g. right side, row 4 of Figure \ref{fig:sup_mat_ssp3d}) and uncertainty in the predicted SMPL shape distribution.

\noindent \textbf{3DPW qualitative results.}
Figure \ref{fig:sup_mat_3dpw} shows qualitative results, particularly focusing on pose prediction, using single-image inputs from 3DPW \cite{vonMarcard2018}. The first column on each side shows the input images. The second column shows the corresponding silhouette and joint heatmap proxy representation predictions. The third column shows the predicted SMPL \cite{SMPL:2015} body (rendered) for each image, obtained by passing the mean of predicted SMPL parameter distributions through the SMPL function. The fourth column visualises the 3D per-vertex uncertainty (or variance) in the SMPL pose and shape distribution predictions (per-vertex uncertainties are mostly due to pose variance rather than shape). 

Specifically, note the large uncertainties of vertices belonging to body parts that are invisible in the image (and corresponding proxy presentations), either due to occluding objects, self-occlusion or being out-of-frame. Furthermore, large uncertainties also occur when the proxy representation prediction is highly-degraded, such as left side, row 7 of Figure \ref{fig:sup_mat_3dpw}.

\clearpage

{\small
\bibliographystyle{ieee_fullname}
\bibliography{egbib}
}

\end{document}